\documentclass[10pt,twocolumn,letterpaper]{article}

\usepackage{iccv}
\usepackage{times}
\usepackage{epsfig}
\usepackage{graphicx}
\usepackage{amsmath}
\usepackage{amssymb}
\usepackage{booktabs}
\usepackage{array}
\usepackage{diagbox}
\usepackage{multirow}
\usepackage[ruled,linesnumbered]{algorithm2e}
\usepackage{paralist}
\usepackage{multicol}
\usepackage[accsupp]{axessibility}
\usepackage{makecell}

\usepackage[pagebackref=true,breaklinks=true,letterpaper=true,colorlinks,bookmarks=false]{hyperref}

\newcolumntype{L}[1]{>{\raggedright\arraybackslash}p{#1}}
\newcolumntype{C}[1]{>{\centering\arraybackslash}p{#1}}
\newcolumntype{R}[1]{>{\raggedleft\arraybackslash}p{#1}}

\iccvfinalcopy 

\def\iccvPaperID{11280} 

\newcommand{\approach}{parse-then-place}

\begin{document}

\title{A Parse-Then-Place Approach for Generating Graphic Layouts \\ from Textual Descriptions}

\author{
    Jiawei Lin\textsuperscript{1}\thanks{Work done during an internship at Microsoft Research Asia.},
    Jiaqi Guo\textsuperscript{2},
    Shizhao Sun\textsuperscript{2},
    Weijiang Xu\textsuperscript{2},
    Ting Liu\textsuperscript{1}, \\
    Jian-Guang Lou\textsuperscript{2},
    Dongmei Zhang\textsuperscript{2} \\
    \textsuperscript{1}Xi’an Jiaotong University, \textsuperscript{2}Microsoft Research Asia \\
    \texttt{\small kylelin@stu.xjtu.edu.cn, tingliu@mail.xjtu.edu.cn,} \\
    \texttt{\small \{jiaqiguo, shizsu, weijiangxu, jlou, dongmeiz\}@microsoft.com}
}

\maketitle
\ificcvfinal\thispagestyle{empty}\fi

\begin{abstract}
Creating layouts is a fundamental step in graphic design.
In this work, we propose to use text as the guidance to create graphic layouts, i.e., \textbf{Text-to-Layout}, aiming to lower the design barriers.
Text-to-Layout is a challenging task, because it needs to consider the implicit, combined, and incomplete layout constraints from text, each of which has not been studied in previous work.
To address this, we present a two-stage approach, named \textbf{\approach{}}.
The approach introduces an intermediate representation (IR) between text and layout to represent diverse layout constraints.
With IR, Text-to-Layout is decomposed into a parse stage and a place stage.
The parse stage takes a textual description as input and generates an IR, in which the implicit constraints from the text are transformed into explicit ones.
The place stage generates layouts based on the IR.
To model combined and incomplete constraints, we use a Transformer-based layout generation model and carefully design a way to represent constraints and layouts as sequences.
Besides, we adopt the pretrain-then-finetune strategy to boost the performance of the layout generation model with large-scale unlabeled layouts.
To evaluate our approach, we construct two Text-to-Layout datasets and conduct experiments on them.
Quantitative results, qualitative analysis, and user studies demonstrate our approach’s effectiveness.
\end{abstract}

\section{Introduction}
\label{sec:intro}

\begin{figure}[htbp]
    \centering
    \includegraphics[width=1.0\linewidth]{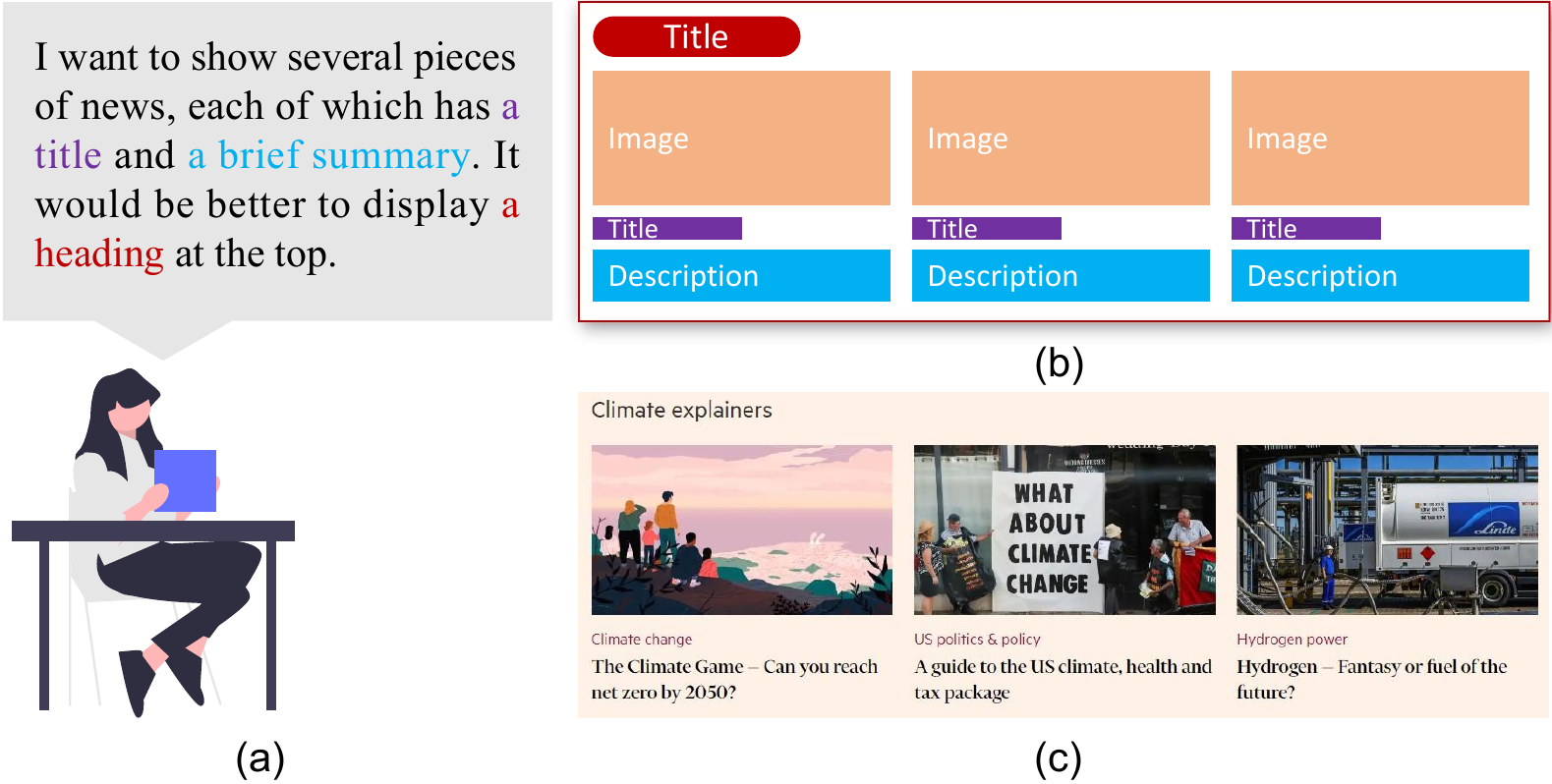}
    \caption{An example of Text-to-Layout. (a) A user describes the desired layout in natural language. (b) A visually appealing layout is automatically generated. (c) The layout is a fundamental ingredient of the final graphic design.}
    \label{fig:example}
\end{figure}




Graphic design is ubiquitous in our daily life.
\emph{Layout}, the sizes and positions of design elements, is fundamental to a graphic design.
However, creating layouts is a complex task that requires design expertise and consumes much time.
While we may not have design knowledge and be unfamiliar with design tools, we have a strong linguistic ability to express our requirements.
That is also what we are doing in the communication with designers.
Thus, we propose to use \emph{text} as the guidance to create graphic layouts, i.e., \emph{Text-to-Layout} (see Figure~\ref{fig:example}).
This enables people without expertise to participate in the design.
Moreover, it helps professional designers create drafts more efficiently, thereby reducing their workloads.
Besides, it also makes the discussion between users and designers much smoother.


While automatic layout generation has been intensively studied~\cite{li2019layoutgan,arroyo2021variational,gupta2021layouttransformer,kong2021blt}, Text-to-Layout is rarely investigated and remains a challenging task due to its unique way to specify layout constraints.
\emph{(i) Implicit Constraint}.
Textual description tends to be abstract and vague, and thus layout constraints from the text are often specified in an implicit and vague way.
Consider the description in Figure~\ref{fig:example}.
It describes the elements in a layout by their functional roles (e.g., news title and summary) rather than their primitive element types (e.g., text box).
It also states a vague, subjective position constraint (e.g., heading at the top) and implicitly poses a hierarchy constraint (e.g., news should be organized in a list).
This characteristic makes Text-to-Layout drastically different from other conditional layout generation tasks where constraints are explicitly specified~\cite{attribute-condition,lee2020ndn,kong2021blt}.
\emph{(ii) Combined Constraint}.
Various types of layout constraints are often jointly specified in text.
For example, the description in Figure~\ref{fig:example} poses 4 different kinds of constraints in total, including element type constraint (e.g., news title), size constraint (e.g., brief summary), position (e.g., heading at the top) and hierarchy (e.g., all news arranged in a list).
However, existing conditional layout generation approaches only tackle one certain kind of constraint at a time.
Hence, how to model the combined constraints and create visually appealing layouts that satisfy all constraints simultaneously remains to be explored.
\emph{(iii) Incomplete  Constraint}.
Users tend not to describe all the elements in a layout, because doing so is extremely tedious.
For instance, the description in Figure~\ref{fig:example} does not specify the image in each piece of news, but the images are indispensable for engaging audiences' attention.
Thus, it is necessary to auto-complete the omitted yet important elements.
While previous work~\cite{gupta2021layouttransformer,jiang2022unilayout} has studied the task of layout completion, it has not been jointly considered with other conditional layout generation tasks.
In addition to the above challenges, Text-to-Layout also faces the serious problem of scarcity of labeled data.
Unlike the Text-to-Image task that has billions of text-image pairs from the Internet~\cite{ramesh2021dalle1,ramesh2022dalle2,saharia2022imagen}, it is prohibitively expensive to collect a similarly sized dataset for Text-to-Layout.

To tackle the challenging Text-to-Layout, our intuition is three-fold.
First, implicit layout constraints in a text could be transformed into explicit ones.
Considering the description in Figure~\ref{fig:example}, the functional roles of elements can be transformed to corresponding primitive element type constraints.
Since this transformation does not require any layout generation capability, we can take advantage of pretrained language models (PLM) that have achieved impressive natural language understanding performance even in a low-data regime~\cite{brown2020language,raffel2020t5,schucher-etal-2022-power}.
In addition, generating layouts conditioned on explicit constraints is a well-studied setting in prior work~\cite{arroyo2021variational,kong2021blt,jiang2022unilayout}.
We can learn from their successful experience to address the problem.
Second, the state-of-the-art approach~\cite{jiang2022unilayout} has formulated conditional layout generation as a sequence-to-sequence transformation problem and demonstrated the superiority of representing constraints and layouts as sequences.
This motivates us to take the same formulation and seek a way to represent combined and incomplete constraints in Text-to-Layout as sequences.
Third, graphic designs are ubiquitous and their layouts are often abundantly available~\cite{bai2021uibert,lee2022pix2struct}.
Though these layouts do not have corresponding textual descriptions, their large quantity and rich layout patterns are highly valuable for learning to generate high-quality, diverse layouts, especially when only scarce labeled data is available.

\begin{figure*}[t!]
    \centering
    \includegraphics[width=\linewidth]{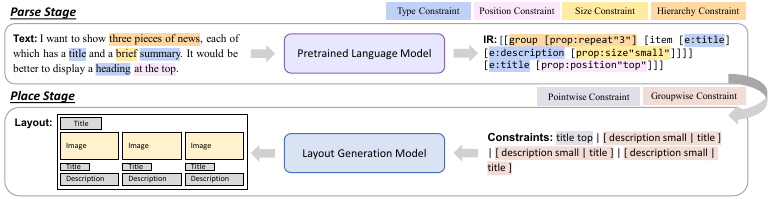}
    \caption{An illustration of our approach. We decompose Text-to-Layout into a parse stage and a place stage. The parse stage maps a textual description into an intermediate representation (IR), in which implicit and vague constraints are transformed into explicit ones. The place stage generates visually pleasing layouts according to the combined constraints. Meanwhile, the place stage completes reasonable elements (the three \emph{image}s in the generated layout) automatically. 
    }
    \vspace{-0.2mm}
    \label{fig:pipeline}
\end{figure*}

Motivated by the intuitions, we propose a two-stage approach for Text-to-Layout, called \emph{\approach{}} (see Figure~\ref{fig:pipeline}).
The approach introduces an \emph{intermediate representation} (IR) between text and layout to formally represent diverse layout constraints, such as element type, size, and hierarchy.
By introducing IR, our approach decomposes Text-to-Layout into two stages: \emph{parse} and \emph{place}.
The \emph{parse} stage takes a textual description as input and outputs IR.
Since IR is a representation of layout constraints specified in the text, we formulate the parse stage as a natural language understanding problem and finetune the T5~\cite{raffel2020t5} PLM to map the text to IR.
The \emph{place} stage generates layouts according to the constraints stated in IR.
Inspired by UniLayout~\cite{jiang2022unilayout}, we use a Transformer-based layout generation model and carefully design an input-output sequence format to represent combined, incomplete constraints and layouts.
In addition, owing to the two-stage design of our approach, we can leverage large-scale unlabeled layouts to pretrain the layout generation model and then finetune it with labeled data.

To evaluate our approach, we construct two Text-to-Layout datasets: \emph{Web5K} and \emph{RICO2.5K}.
Web5K targets Web page layouts and contains 4,790 $\langle \texttt{text}, \texttt{IR}, \texttt{layout}\rangle$ samples, while RICO2.5K targets Android UI layouts and includes 2,412 samples.
The quantitative and qualitative results on both datasets show that \approach{} significantly outperforms baseline approaches in terms of perceptual quality and consistency.
We also conduct a user study to evaluate our approach more comprehensively.
Compared to the baseline approaches, users find that our generated layouts better match textual descriptions in 47.6\% and 56.0\% of trials, and have higher quality in 52.2\% and 62.5\% of trials in Web5K and RICO2.5K, respectively.

\section{Related Work}
\label{sec:related work}

\noindent\textbf{Graphic layout generation.}
The automatic generation of aesthetic layouts has fueled growing interest.
Early work in this field primarily studies unconditional layout generation using advanced generative models, such as Generative adversarial Network (GAN)~\cite{li2019layoutgan} and Variational Autoencoder (VAE)~\cite{arroyo2021variational, Jiang2022Coarse, Yamaguchi2021ICCV}.
To enable more practical usages, recent work explores conditional layout generation, where the goal is to generate layouts conforming to certain constraints.
Li et al.~\cite{attribute-condition} introduce a GAN-based approach to incorporate the meta attributes of elements, e.g., reading order, expected area, and aspect ratio, for layout generation.
Jyothi et al.~\cite{jyothi2019layoutvae} present LayoutVAE, a two-stage VAE-based framework that generates layouts given a set of element types.
Lee et al.~\cite{lee2020ndn} propose Neural Design Network, a three-stage VAE-based method that generates layouts based on a set of elements and their partial geometric relations.
Gupta et al.~\cite{gupta2021layouttransformer} propose LayoutTransformer for layout completion.
Given an initial layout with only one element, LayoutTransformer leverages self-attention to complete it autoregressively.
Kong et al.~\cite{kong2021blt} observe that LayoutTransformer's pre-defined generation order hinders it from performing tasks where the constraints disagree with the pre-defined order.
To alleviate this issue, they report Bidirectional Layout Transformer to generate layouts in a non-autoregressive manner.
The above work adopts customized model architectures and optimization methods for different conditional layout generation tasks.
Jiang et al.~\cite{jiang2022unilayout} instead propose UniLayout, which formulates conditional layout generation as a sequence-to-sequence transformation problem and adopts the Transformer encoder-decoder architecture to address it.
They further design an input sequence format to represent diverse constraints in a unified way so that the model can handle various tasks.
Albeit effective, these methods cannot be directly adopted to Text-to-Layout.
On the one hand, they all require explicit constraints as input, but the constraints from text are implicit and vague.
On the other hand, they only tackle one certain kind of constraint at a time, but the constraints from text are often combined and incomplete, requiring an ability to take into consideration various kinds of constraints.
To bridge this gap, our approach's parse stage transforms implicit constraints in the text to explicit ones, and the place stage extends UniLayout to enable layout generation conditioned on combined and incomplete constraints.

\noindent\textbf{Text-to-Layout in other scenarios.}
There has been several attempts at using text as the guidance to create other types of layouts.
House Plan Generative Model (HPGM) is a pioneering work on generating building layouts from textual descriptions~\cite{3d-house}.
Given a description, HPGM first parses it into a structural graph representation and then predicts room attributes according to the graph.
Tan et al.~\cite{text2scene2019} study scene layout generation from text and propose Text2Scene, an end-to-end approach that sequentially generates objects and their attributes.
Radevski et al.~\cite{radevski-etal-2020-decoding} target the same problem and present SR-BERT.
It is built upon BERT~\cite{devlin-etal-2019-bert}, and it generates layouts in an iterative and non-autoregressive manner.
Hong et al.~\cite{Hong2018Inferring} introduce semantic layout generation from text and regard it as an intermediate step for Text-to-Image.
Similarly, Huang and Canny~\cite{Huang2019Sketchforme} propose to generate sketch layouts from text.
They use the layouts as bottleneck representations for Text-to-Sketch.
These methods are highly customized for their targeted scenarios.
Adapting them to graphic layouts is non-trivial and of inferior performance, as shown in our experiments.


\section{Methodology}
\label{sec:method}

\begin{table*}
\centering
    \begin{small}
    \resizebox{\textwidth}{!}{
        \begin{tabular}{cll}
            \toprule
            \# & Textual Description & Intermediate Representation \\
            \midrule
            1 & a news heading & \texttt{[e:title]} \\ 
            2 & a news heading at the top & \texttt{[e:title [prop:position"top"]]} \\ 
            3 & a brief news summary & \texttt{[e:description [prop:size"small"]]} \\ 
            4 & 3 news pieces. each has a title and summary & \texttt{[group [prop:repeat"3"] [item[e:title][e:description]]]} \\ 
            \bottomrule
        \end{tabular}
        }
    \end{small}
    \caption{Illustrative examples of intermediate representation.}
     \label{Tab:ir_grammar}
\end{table*}

In this section, we elaborate on our~\approach{} approach for generating a layout $y$ from a textual description $x$.
As illustrated in Figure~\ref{fig:pipeline}, our approach introduces an intermediate representation (IR) and decomposes the generation into the following two stages:
\begin{compactenum}
    \item \textbf{Parse Stage}: Translate the textual description $x$ into IR $z$.
    As IR is a formal representation of layout constraints, we formulate this stage as a natural language understanding problem and take advantage of pretrained language models (PLM) to address it.

    \item \textbf{Place Stage}: Generate a layout $y$ conditioned on IR $z$.
    We formulate this stage as a sequence-to-sequence transformation problem and use a Transformer-based layout generation model.
    To support combined and incomplete constraints, we carefully design an input-output sequence format to represent constraints and layouts.
    In addition, we adopt a pretrain-then-finetune strategy to improve the layout generation model with large-scale unlabeled layout data.
\end{compactenum}

In what follows, we first introduce the IR and then elaborate on the parse and place stages, respectively.

\subsection{Intermediate Representation}
\label{sec:ir}

As a bridge between text and layout, IR should meet the following two requirements.
\emph{(i) Expressiveness}.
IR should be expressive to represent diverse user constraints on layouts.
Otherwise, the mapping from text to IR will incur undesired information loss, and consequently, the generated layouts will fail to satisfy user constraints.
\emph{(ii) Formalism}.
IR should have a good formalism so that existing natural language understanding techniques can exhibit accurate understanding performance.
This is because previous studies~\cite{guo-etal-2020-benchmarking-meaning,herzig2021unlocking} show that the formalism of formal languages could significantly impact understanding performance.

To this end, we design an IR based on the above requirements.
Table~\ref{Tab:ir_grammar} shows 4 illustrative examples of our designed IR.
In our preliminary study on Text-to-Layout,\footnote{We invited people without professional graphic design skills to describe a set of layouts and studied their descriptions' characteristics.} 
we found that users tend to describe layouts by specifying the functional roles (Table~\ref{Tab:ir_grammar} \#1), positions (\#2), sizes (\#3), and hierarchies (\#4) of elements.
Hence, our IR currently supports these constraints\footnote{Notably, it is easy to extend the IR for other types of constraints.}.
In terms of formalism, the IR follows the hierarchical representation scheme~\cite{gupta-etal-2018-semantic-parsing}, which has exhibited accurate understanding performance~\cite{Rongali2020Dont}.
More details and examples of the IR can be found in the appendix.

\subsection{Parse Stage}
\label{sec:parse}

The parse stage aims to map a description $x$ to IR $z$, so that implicit constraints in text are transformed into explicit constraints.
We use T5~\cite{raffel2020t5}, a Transformer encoder-decoder based PLM to address this natural language understanding problem.
While T5 is pretrained on text data, it has been shown to be effective at translating natural language to formal language even in a low-data regime~\cite{UnifiedSKG,schucher-etal-2022-power}.
Specifically, the encoder takes a description $x = \left\{x_1,\cdots,x_n\right\}$ as input, where $x_i$ is the $i$-th token in the description, and it outputs the contextualized representations $h^x = \left\{h^x_1,\cdots,h^x_n\right\}$.
Then the decoder autoregressively predicts the tokens of IR $P_\theta\left(z_j|z_{<j}, x\right)$ by attending to previously generated tokens $z_{<j}$ (via self-attention) and the encoder outputs $h^x$ (via cross-attention).
During training, we finetune T5 parameters $\theta$ by minimizing the negative log-likelihood of labeled IR $z$ given input text $x$:
\begin{equation}
\mathcal{L}_\theta = - \frac{1}{|\mathcal{D}|}\sum_{(x, z, y) \in \mathcal{D}}\sum^{|z|}_{j=1} \log P_{\theta}(z_j| z_{<j}, x).
\end{equation}


\subsection{Place Stage}
\label{sec:place}

The goal of the place stage is to generate a layout $y$ according to the combined constraints specified in IR $z$.
Inspired by UniLayout~\cite{jiang2022unilayout}, we formulate this stage as a sequence-to-sequence transformation problem and adopt a Transformer encoder-decoder based layout generation model to solve it.
Given an IR $z$, we first deterministically transform it into a constraint sequence $s = \pi(z)$.
The model then takes the sequence $s$ as input and generates a layout sequence $y$.
The parameters $\phi$ of the model are trained to minimize the negative log-likelihood of the layout sequence $y$ given constraint sequence $s$:
\begin{equation}
\label{equ:place-finetune}
\mathcal{L}_\phi^{\textsc{FT}} = - \frac{1}{|\mathcal{D}|}\sum_{(x, z, y) \in \mathcal{D}} \sum^{|y|}_{j=1} \log P_{\phi}(y_j| y_{<j}, \pi(z)).
\end{equation}
To support combined and incomplete constraints, we carefully design the constraint and layout sequences, as introduced below.

\noindent\textbf{Constraint Sequence.}
To represent combined constraints, we serialize each constraint as a sub-sequence and concatenate all of them in a certain order.

First, we divide layout constraints into the following two categories, which have distinct characteristics and thus require different serialization strategies.
\emph{(i) Pointwise constraint} refers to the constraint specific to a single element.
The element type, position and size constraints in Figure~\ref{fig:pipeline} all fall into this category.
\emph{(ii) Groupwise constraint} denotes the hierarchy between a group of elements.
It includes containment (e.g., a toolbar with two icons and a textbox) and arrangement (e.g., three news lists) relationships between elements.
It is worth noting that the layout constraints studied in prior work (see Section~\ref{sec:related work}) are all within our consideration.


For pointwise constraint, we denote it using a pre-defined constraint token $k \in \Sigma$, where $\Sigma$ is a vocabulary that varies with different constraints (e.g., the vocabulary for position constraint could be $\Sigma_{pos} = \{\text{left}, \cdots, \text{top}\}$).
Multiple pointwise constraints on an element are represented as a concatenation of constraint tokens:
$C^{po} = \left\{k_{type}\ k_{pos}\ k_{size}\right\}$, where $k_{type}$, $k_{pos}$ and $k_{size}$ are element type, position and size constraints, respectively.
For example, we use \underline{image left large} to express ``a large image on the left''.
In terms of groupwise constraint, we take inspiration from recent work that represents structural knowledge bases (e.g., knowledge graph and table) as sequences~\cite{UnifiedSKG,lin-etal-2021-leveraging}, and use brackets to denote groups $C^{gp}$.
For instance, the hierarchy ``a news piece with an image and a title at the top'' can be represented as \underline{[ image $\vert$ title top ]}.
With these serialization strategies, we transform the combined constraints into a sequence by sorting the sub-sequences in the alphabetic order of element types and concatenating them with a separator $|$:
\begin{equation*}
\label{eq:input format}
s = \left\{C^{po}_1|\cdots|C^{po}_{p}|C^{gp}_1|\cdots|C^{gp}_q\right\}.
\end{equation*}
We provide some examples of constraint sequences in the appendix.
Since IR is a formal language, it can be deterministically translated into corresponding constraint sequences.

\noindent\textbf{Layout Sequence.}
To facilitate the model learning to complete omitted elements given incomplete constraints, we introduce a categorical attribute $a \in \left\{\text{complete}, \text{null}\right\}$ for each element to indicate whether it is auto-completed or not.
It helps the model better distinguish the auto-completed elements.
Each element $e_i$ now has 6 attributes, including the newly-introduced attribute $a_i$, element type $c_i$, left coordinate $l_i$, top coordinate $t_i$, width $w_i$ and height $h_i$.
Following previous work~\cite{jiang2022unilayout, gupta2021layouttransformer, kong2021blt}, we represent an element as a sequence with 6 discrete tokens $e_i = \left\{a_i c_i l_i t_i w_i h_i\right\}$, in which the continuous attributes $l_i$, $t_i$, $w_i$ and $h_i$ are discretized into integers between $[0, n_{bins}-1]$.
Then, we represent a layout by sorting all the elements in the alphabetic order and concatenating all their tokens using a separator $|$:
\begin{equation*}
\label{eq:input format_layout_seq}
y = \left\{a_1 c_1 l_1 t_1 w_1 h_1|\cdots|a_m c_m l_m t_m w_m h_m\right\}.
\end{equation*}

\noindent\textbf{Exploiting Unlabeled Layouts.}
Learning to generate high-quality, diverse layouts from combined, incomplete constraints is challenging, especially when only a small volume of labeled data is available.
But fortunately, graphic layouts are easily accessible in the wild.
Though they do not have corresponding textual descriptions, their wealthy layout patterns are tremendously helpful for acquiring layout generation skills. 
Therefore, we attempt to pretrain the layout generation model with large-scale unlabeled layouts, and then finetune it with labeled data.

For pretraining, we curate a synthetic dataset $\mathcal{D}_s = \left\{(\hat{z}_i, y_i)\right\}^{|\mathcal{D}_{s}|}_i$ from unlabeled layouts, where IR $\hat{z}_i$ is synthesized from layout $y_i$, and we use it to pretrain the model:
\begin{equation}
\mathcal{L}_\phi^{\textsc{PT}} = - \frac{1}{|\mathcal{D}_s|}\sum_{(\hat{z}, y) \in \mathcal{D}_s} \sum^{|y|}_{j=1} \log P_{\phi}(y_j| y_{<j}, \pi{(\hat{z})}).
\end{equation}
This dataset is built upon the layout's structural nature, making it feasible to extract desired constraints from unlabeled layouts through some heuristic rules.
For example, we can infer position constraints from the position property of elements in the layout source code.
In addition, the hierarchical characteristics of source code also facilitate the extraction of hierarchy constraints.
With these constraints, we can synthesize an IR from a layout and then build the synthetic dataset.
The details of IR synthesis are given in the appendix.

\begin{table*}[t]
    \centering
    \begin{small}
     \resizebox{\textwidth}{!}{
        \begin{tabular}{lcccccccc}
        \toprule
         & \multicolumn{8}{c}{WebUI}  \\
         \cmidrule(l){2-9} 
         Method & FID $\downarrow$ &  Align. $\downarrow$ & Overlap $\downarrow$ & mIoU $\uparrow$ & UM $\uparrow$ & Type Cons. $\uparrow$ & Pos \& Size Cons. $\uparrow$ & Hierarchy Cons. $\uparrow$ \\
         \midrule
        Mockup~\cite{mockup} & 37.0123 & 0.0059 & 0.4348 & 0.1927 & 0.4299 & 0.6851 & 0.5508 & - \\
        Text2Scene~\cite{text2scene2019} & 27.1612 & 0.0042 & 0.4455 & 0.1899 & 0.4164 & 0.7515 & 0.5377 & - \\
        SR-BERT~\cite{radevski-etal-2020-decoding} & 10.1640 & 0.0032 & 0.6501 & 0.2322 & 0.4127 & 0.8551 & 0.6423 & - \\
        Ours & \textbf{2.9592} & \textbf{0.0008} & \textbf{0.1380} & \textbf{0.6841} & \textbf{0.5080} & \textbf{0.8864} & \textbf{0.8086} & \textbf{0.4622} \\
        \midrule
        Real data & - & 0.0007 & 0.1343 & - & - & - & - & - \\
        \bottomrule
        \end{tabular}
        }
        \resizebox{\textwidth}{!}{
        \begin{tabular}{lcccccccc}
        \toprule
         & \multicolumn{8}{c}{RICO}  \\
         \cmidrule(l){2-9} 
         Method & FID $\downarrow$ &  Align. $\downarrow$ & Overlap $\downarrow$ & mIoU $\uparrow$ & UM $\uparrow$ & Type Cons. $\uparrow$ & Pos \& Size Cons. $\uparrow$ & Hierarchy Cons. $\uparrow$ \\
        \midrule 
        Mockup~\cite{mockup} & 29.5170 & 0.0096 & 0.5416 & 0.1868 & 0.2892 & 0.7548 & 0.5724 & - \\
        Text2Scene~\cite{text2scene2019} & 23.4324 & 0.0072 & 0.4558 & 0.1972 & 0.3512 & 0.8093 & 0.6061 & - \\
        SR-BERT~\cite{radevski-etal-2020-decoding} & 13.1268 & 0.0055 & 0.6373 & 0.2927 & 0.3306 & 0.9295 & 0.6425 & - \\
        Ours & \textbf{4.9256} & \textbf{0.0015} & \textbf{0.2918} & \textbf{0.5267} & \textbf{0.3661} & \textbf{0.9539} & \textbf{0.7145} & \textbf{0.7841} \\
        \midrule
        Real data & - & 0.0029 & 0.2714 & - & - & - & - & - \\
        \bottomrule
        \end{tabular}
        }
    \end{small}
    \caption{Quantitative results. The \emph{perceptual quality} of generated layouts is measured in columns 2-6. The \emph{consistency} between the generated layout and the description is measured in columns 7-9, including Type Cons., Pos \& Size Cons. and Hierarchy Cons.
    }
    \vspace{-.4cm}
    \label{tab:Quantitative}
\end{table*}

After pretraining, we finetune the model with labeled data (see Equation~\ref{equ:place-finetune}) to close the distribution gap between synthetic and labeled data.
There are two main reasons for this gap.
First, the characteristics of elements that are commonly omitted by users cannot be perfectly described in the synthesized IR.
Second, position and size constraints are somewhat subjective, so there is an inevitable gap between the constraints extracted by rules and human perception.


\subsection{Inference}
At inference time, our approach can generate multiple layouts for a textual description.
The parse stage takes as input a description and generates an IR with the largest likelihood (\textit{argmax sampling}).
The place stage deterministically transforms the IR into a constraint sequence and generates multiple layouts via \textit{Top-K sampling}, which samples the next token from the top k most probable choices~\cite{fan-etal-2018-hierarchical}.

\section{Experiments}

\subsection{Setups}
\noindent\textbf{Datasets.}
We conduct experiments on two graphic layout datasets, including \textbf{RICO}~\cite{deka2017rico} and \textbf{WebUI}.
RICO is a public dataset of Android UI layouts without textual descriptions.
As introduced in Section~\ref{sec:method}, our method uses $\langle \texttt{text}, \texttt{IR}\rangle$ labeled data during the parse stage.
In the place stage, it uses unlabeled \texttt{layout} data for pretraining and $\langle \texttt{IR}, \texttt{layout}\rangle$ labeled data for finetuning.
Thus, we labeled 2,412 $\langle \texttt{text}, \texttt{IR}, \texttt{layout}\rangle$ triplets (denoted as \emph{RICO2.5K}) for the RICO dataset, and left the other 40k layouts in the dataset as unlabeled data (denoted as \emph{RICO40K}).
WebUI is a dataset of web UI layouts crawled from the Internet by ourselves.
Similarly, we labeled 4,790 data (denoted as \emph{Web5K}) and left 1.5 million layouts as unlabeled data (denoted as \emph{Web1.5M}).
Moreover, we split the labeled data (i.e., RICO2.5K and Web5K) into training, validation and test sets by ratios of $80\%$, $10\%$ and $10\%$. 

Specifically, to ensure the coverage and quality of the labeled datasets, we create them by following steps: 1) sampling from the original unlabeled set, 2) training annotators, 3) labeling the data by annotators, and 4) examining annotation quality by experts.
Please see the appendix for examples of labeled data and details of the annotation process.

\noindent\textbf{Baselines.}
While there is no existing work in our scenario (i.e., graphic layouts), there are strong methods in relevant scenarios (e.g., scenes).
We adapt them as our baselines.
\textbf{MockUp}~\cite{mockup} encodes a description with BERT~\cite{devlin-etal-2019-bert} and adopts Transformer decoder to generate a layout autoregressively.
\textbf{Text2Scene}~\cite{text2scene2019} leverages an RNN-based text encoder to encode a description, and recursively predicts the next element in a layout and its corresponding bounding box by the proposed convolutional recurrent module.
\textbf{SR-BERT}~\cite{radevski-etal-2020-decoding} first obtains a description's contextual representations from BERT, and then generates a layout in an iterative and non-autoregressive manner.
They all require the description and layout pairs as the training data.

\noindent\textbf{Evaluation Metrics.}
We evaluate generation performance from two aspects.
The first aspect is the \textbf{perceptual quality}, reflecting whether the generated layout looks aesthetically pleasing and diverse.
Following previous work, we consider five metrics.
\emph{Fr\'{e}chet Inception Distance (FID})~\cite{kikuchi2021constrained} measures the distance between the distribution of generated layouts and that of real layouts, which reflects both aesthetic quality and diversity.
\emph{Alignment (Align.})~\cite{attribute-condition} indicates whether elements in the generated layout are well-aligned.
\emph{Overlap}~\cite{attribute-condition} measures the overlap area between two arbitrary elements in the generated layout.
\emph{Maximum IoU (mIoU})~\cite{kikuchi2021constrained} measures the maximum IoU between the elements in the generated layout and those in the real layout.
Align., Overlap and mIoU mainly measure aesthetic quality.
\emph{Unique Match (UM})~\cite{arroyo2021variational} uses DocSim~\cite{patil2020read} to retrieve the most similar layout from the training set for each generated layout and then computes the ratio of the number of distinct retrieved layouts to the total number of generated layouts. 
It mainly indicates diversity.

The second aspect is the \textbf{consistency}, reflecting whether the generated layout matches the description.
We measure it by the satisfaction rate of constraints~\cite{lee2020ndn}.
Specifically, we subdivide the constraints into three groups based on the discussion in Section~\ref{sec:ir} and compute the satisfaction rate for each of them, denoted as \emph{Type Consistency}, \emph{Position \& Size Consistency} and \emph{Hierarchy Consistency}\footnote{Hierarchy Consistency is only computed in our method since hierarchies are hard to parse from the layouts generated by baseline methods.} respectively.

\begin{figure*}
    \centering
    \includegraphics[width=\linewidth]{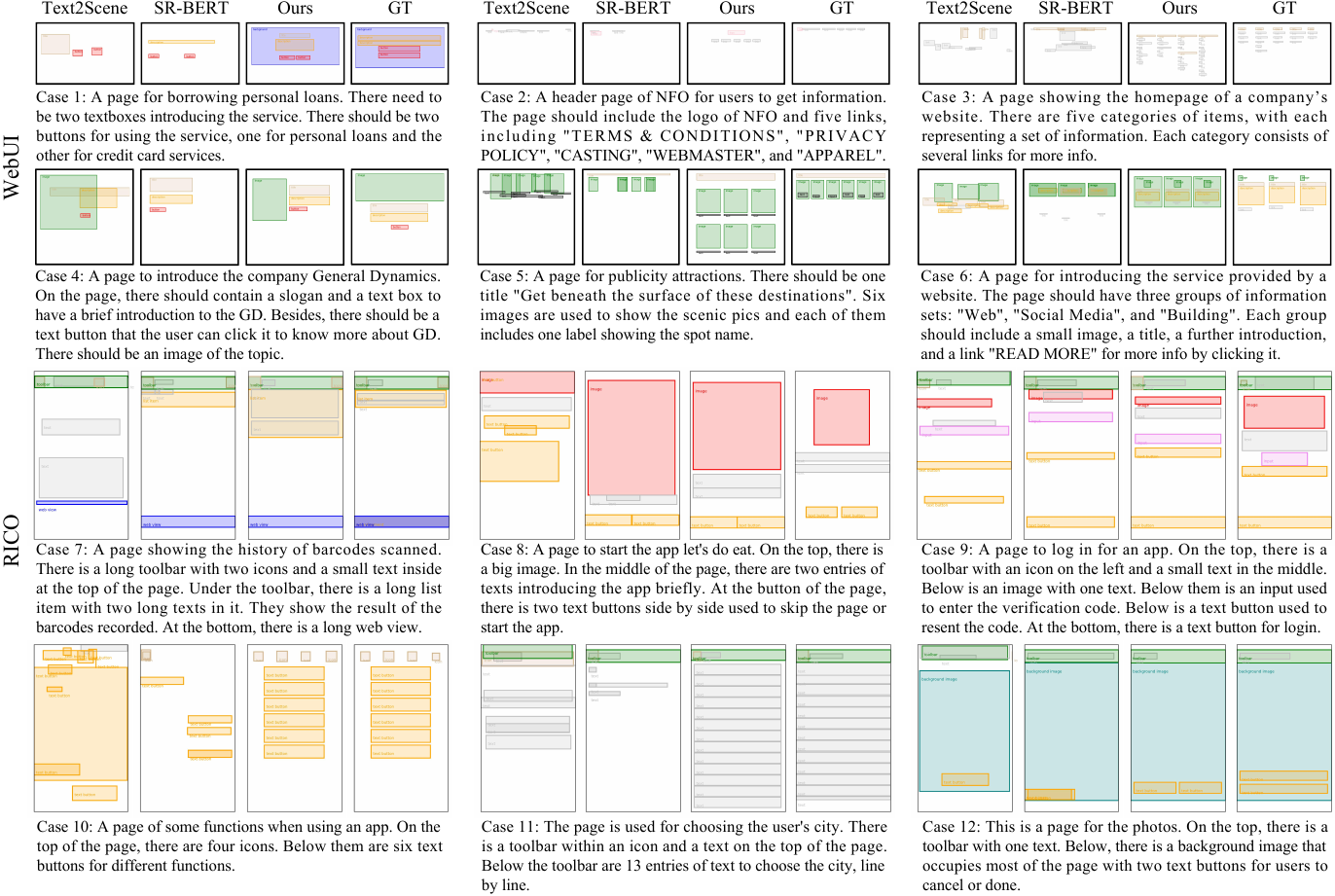}
    \caption{Qualitative comparison with the best two baselines (zoom-in for better view). The real layout corresponding to the description is denoted as GT. The generated layout does not have to be the same as GT, as long as it is visually pleasing and matches the description.
    }
    \label{fig:qualitative_result}
    \vspace{-.2cm}
\end{figure*}

\noindent\textbf{Implementation Details.}
Our approach is implemented with PyTorch~\cite{PyTorch} and Huggingface~\cite{wolf-etal-2020-transformers}.
In the parse stage, we use T5-base.
In the place stage, the layout generation model has 12 encoder layers and 12 decoder layers.
The hidden dimension is set to 768, and the number of attention heads is set to 8.
All the models are optimized with Adam optimizer~\cite{kingma2014adam} on NVIDIA V100 GPUs.
The layouts in WebUI and RICO are proportionally scaled to $120\times120$ and $144\times256$ respectively.
See the appendix for more details.

\subsection{Main Results}
\noindent\textbf{Quantitative Analysis.}
Table~\ref{tab:Quantitative} shows the quantitative results.
Our method significantly outperforms the baselines on all metrics, indicating that it can generate layouts that are more visually pleasing and consistent with the descriptions.
Moreover, our method can generate layouts with similar or even better Align. and Overlap compared to real data.

\begin{figure*}
    \centering
    \includegraphics[width=\linewidth]{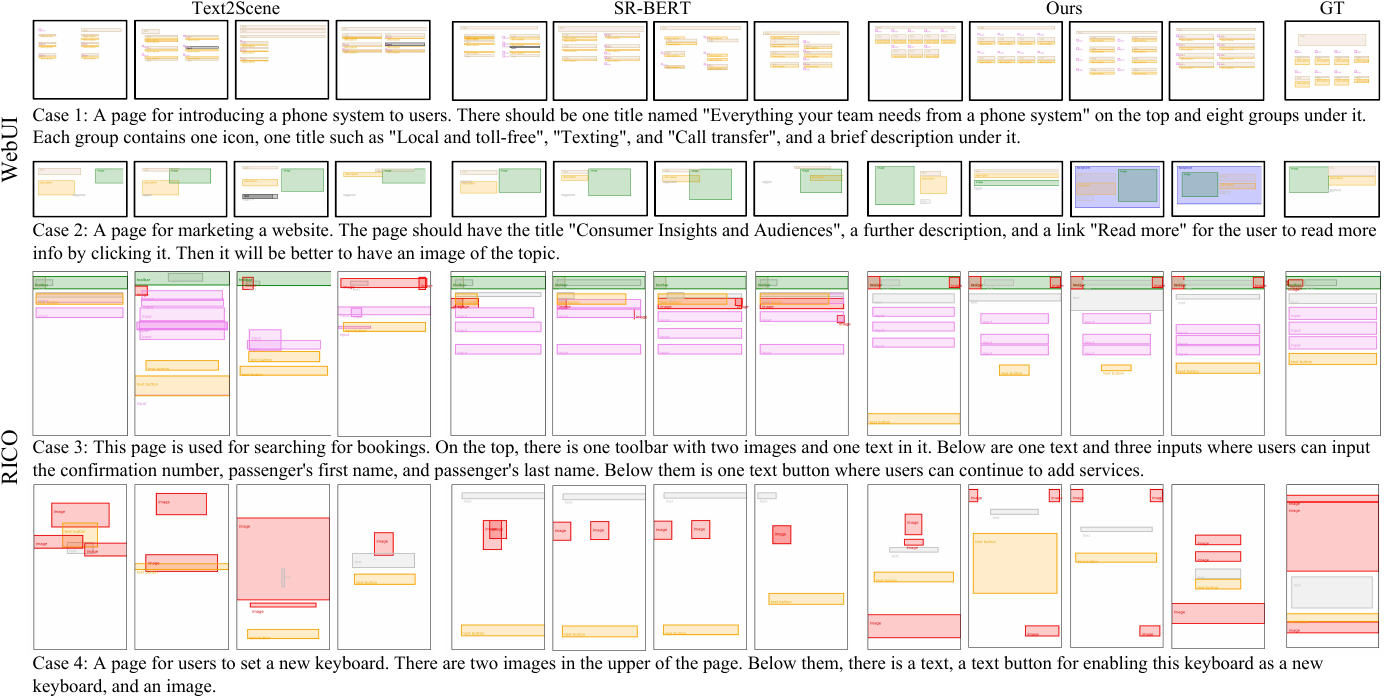}
    \caption{Qualitative comparison of layout diversity with the best two baselines. 
    We show four layouts for each description.
    }
    \label{fig:qualitative_result_multi}
\end{figure*}

\noindent\textbf{Qualitative Analysis.}
Figure~\ref{fig:qualitative_result} and Figure~\ref{fig:qualitative_result_multi} show qualitative results.
The results demonstrate that the layouts from our approach are of high quality, while the layouts from baselines frequently contain misalignment, incorrect overlap, and weird spacing.
Besides, the layouts from our approach are more consistent with the descriptions.
For example, in case 2 of Figure~\ref{fig:qualitative_result}, the baselines fail to generate five links, while our approach succeeds.
Our approach can also complete omitted yet important elements.
For example, in case 1 of Figure~\ref{fig:qualitative_result}, it completes a background image, which is common on web pages.
Figure~\ref{fig:qualitative_result_multi} shows qualitative results for layout diversity, where each method generates four layouts for each description.
The results indicate that our approach can generate diverse layouts with rich patterns and high quality, while other methods either suffer from unsatisfactory quality or homogeneous layout designs.
For example, in case 1 of Figure ~\ref{fig:qualitative_result_multi}, the eight groups are arranged in different ways by our approach.
However, in SR-BERT, they are all arranged in a two-column manner with misalignment and incorrect overlap.

\begin{table}[t]
    \centering
    \begin{small}
    \resizebox{\linewidth}{!}{
    \begin{tabular}{lccc}
    \toprule
      Dataset  & Method & Perceptual Quality $\uparrow$ & Consistency $\uparrow$  \\
      \midrule
      \multirow{3}{*}{WebUI} & Text2Scene & 0.226 & 0.250 \\ 
      & SR-BERT & 0.252 & 0.274 \\
      & Ours & \textbf{0.522} & \textbf{0.476} \\
      \midrule
      \multirow{3}{*}{RICO} & Text2Scene & 0.170 & 0.200 \\
      & SR-BERT & 0.205 & 0.240 \\
      & Ours & \textbf{0.625} & \textbf{0.560} \\
    \bottomrule
    \end{tabular}
    }
    \end{small}
    \caption{Results of user study. For each model, we count how many people prefer the layouts it generates.}
    \vspace{-.4cm}
    \label{tab:user study}
\end{table}

\noindent\textbf{User Study.}
In the user study, we compare with the best two baselines (measured by quantitative metrics), SR-BERT and Text2Scene.
For each method, we randomly sample 500 and 200 generated layouts on WebUI and RICO test sets, respectively.
Then, we invite 7 participants to the study.
Each participant is asked to evaluate 100 groups, each group containing a description and three layouts generated by different models.
They need to answer two questions: 1) which layout has the best perceptual quality, and 2) which layout is the most consistent with the description.
Table~\ref{tab:user study} shows the results.
Our approach surpasses the baselines in both perceptual quality and consistency.


\begin{table}[t]
    \centering
    \begin{small}
    \begin{tabular}{lcc}
    \toprule
      Dataset  & Random Initialization & T5 Initialization \\
      \midrule
      WebUI & 35.5\% & 76.2\% \\
      RICO & 9.9\% & 56.6\% \\
    \bottomrule
    \end{tabular}
    \end{small}
    \vspace{1mm}
    \caption{IR accuracy in the parse stage.}
    \label{tab:ir_accuracy}
    \vspace{-.5cm}
\end{table}

\subsection{Ablation Studies}
Our parse stage utilizes a pretrained language model.
To understand its effect, we compare the performance of the network initialized by T5 pretrained weights and the same network initialized randomly. 
To avoid the influence of the place stage, we only evaluate the performance of the parse stage.
We use IR accuracy as the metric, where a predicted IR is considered correct if all its constraints are the same as those of the golden IR.
Table~\ref{tab:ir_accuracy} shows the results.
On both datasets, IR accuracy improves significantly with T5 pretrained weights initialization.
This suggests that PLM offers great help in transforming implicit constraints in the text into explicit ones.

\begin{table*}[t]
    \centering
    \begin{small}
     \resizebox{\textwidth}{!}{
        \begin{tabular}{lcccccccc}
        \toprule
         & FID $\downarrow$ &  Align. $\downarrow$ & Overlap $\downarrow$ & mIoU $\uparrow$ & UM $\uparrow$ & Type Cons. $\uparrow$ & Pos \& Size Cons. $\uparrow$ & Hierarchy Cons. $\uparrow$ \\
        \midrule
        0 (\emph{w/o pretrain}) & 3.8141 & 0.0010 & 0.2453 & 0.5773 & 0.6304 & 0.8727 & 0.7737 & 0.4447 \\
         50K & 3.3341 & 0.0009 & 0.1792 & 0.6623 & 0.5191 & 0.8863 & 0.8114 & 0.4155 \\
         250K & 3.2846 & 0.0009 & 0.1597 & 0.6783 & 0.5055 & 0.8844 & 0.8229 & 0.4480 \\
         500K & 3.1259 & 0.0009 & 0.1400 & 0.6843 & 0.4998 & 0.8780 & 0.8086 & 0.4291 \\
         1M & 2.9408 & 0.0008 & 0.1504 & 0.6836 & 0.4776 & 0.8837 & 0.8103 & 0.4339 \\
         \midrule
         1.5M (Ours) & 2.9592 & 0.0008 & 0.1380 & 0.6841 & 0.5080 & 0.8864 & 0.8086 & 0.4622 \\
         \emph{\quad w/o two stages} & 4.3163 & 0.0010 & 0.3010 & 0.3168 & 0.5236 & 0.8620 & 0.7691 & 0.4350 \\
         \emph{\quad w/ golden IR} & 2.6555 & 0.0008 & 0.1301 & 0.6866 & 0.5047 & 0.9532 & 0.8286 & 0.5058 \\
         \bottomrule
        \end{tabular}
        }
        \end{small}
    \caption{Ablation studies on WebUI.}
    \label{tab:ablation_place}
    \vspace{-.5cm}
\end{table*}

To study the effect of unlabeled data used in the place stage, we conduct experiments on various volumes of unlabeled layouts: 0, 50K, 250K, 500K, 1M and 1.5M (ours).
Table~\ref{tab:ablation_place} shows the results.
Without using any unlabeled layout, our approach already outperforms all baselines (see Table~\ref{tab:Quantitative}) on every metric, suggesting that our overall algorithm design is very effective.
When more unlabeled layouts are used, the performance is further boosted, especially on perceptual quality metrics (such as FID, Overlap and mIoU), indicating that the rich patterns of unlabeled layouts indeed improve generation skills.

Moreover, since we represent the layouts as sequences, it is feasible to train an end-to-end model using the description and layout pairs.
Specifically, we finetune T5 on Web5K (denoted as \textit{w/o two stages}).
The results show that the end-to-end variant performs worse than our method under the same condition (w/o pretrain), again confirming the effectiveness of our method.
Finally, we input the golden IR into the place stage (denoted as \emph{w/ golden IR}) to study the effect of the parse stage on the overall approach performance.
We see that almost all metrics are significantly improved in this setting.
This leads us to believe that utilizing more powerful PLM can further boost the performance of our approach.

\section{Conclusion}

In this work, we propose to use text as guidance to create graphic layouts, i.e.,  Text-to-Layout.
To tackle this challenging task, we present a two-stage approach, parse-then-place.
It introduces an intermediate representation between text and layout to decompose the problem into a parse and place stage.
Experiments demonstrate the effectiveness of our approach.
However, our approach still has some limitations.
First, it does not perform as well on some high-level descriptions (e.g., a layout exhibiting products) as on detailed descriptions in the datasets.
Second, it currently does not support arbitrarily nested hierarchy constraints (e.g., a modal with three list items, each containing two texts). 
In the future, we plan to generalize our approach to more types of graphic design.
We will also investigate how to extend our intermediate representation to support user constraints in other modalities, e.g., images.

\clearpage
{\small
\bibliographystyle{ieee_fullname}
\bibliography{egbib}
}

\clearpage



\appendix
\onecolumn
\section*{\centering\bfseries\Large Supplementary Materials for \\ A Parse-Then-Place Approach for Generating Graphic Layouts from Textual Descriptions}

\vspace{20mm}

\section{Intermediate Representation}

In our approach, an intermediate representation (IR) is introduced to formally represent the layout constraints, including element type constraints, position constraints, size constraints and hierarchy constraints.
The context-free grammar of our designed IR is presented in Equation~\ref{eq:ir_grammar}.
Each nonterminal symbol, which is to the left of $\xrightarrow{}$, corresponds to a production rule. 
The terminal symbols all start with a lowercase letter, including the keywords (\emph{group}, \emph{position}) and the values (\emph{image}, \emph{top}). 

Specifically, we use $Pos$ and $Size$ to denote position and size constraints.
$Element$ represents a graphic element, containing a $Type$ (required) and several $Prop$s (not necessary).
$Group$ stands for a hierarchy constraint and consists of multiple $Element$s.
Moreover, we use $Num$ to simplify the representation when there are too many elements of the same type.

\begin{equation}
\begin{split}
R &\xrightarrow{} [ A ]\ \vert \ [ AA ] \ \vert \ [ AAA ] \ \vert \ [ AAA \dots A ]  \\
A &\xrightarrow{} Element \ \vert \ Group \\
Prop &\xrightarrow{} Pos \ \vert \ Size \ \vert \ Num \\
Element &\xrightarrow{} [e\colon Type] \ \vert \ [e\colon Type\ Prop] \ \vert \ [e\colon Type\ Prop\ Prop] \ \vert \ [e\colon Type\ Prop\ Prop\ Prop] \\
Group &\xrightarrow{} [ group\ Num [ item\ Element \dots Element ] ] \\
Pos &\xrightarrow{} [prop\colon position\ Pvalue] \\
Size &\xrightarrow{} [prop\colon size\ Svalue] \\
Num &\xrightarrow{} [prop\colon repeat\ Nvalue] \\
 \\
Type &\xrightarrow{} image \ \vert \ text \ \vert \ title \ \vert \ icon \ \vert \ \dots \\
Pvalue &\xrightarrow{} top \ \vert \ bottom \ \vert \ left \ \vert \ right \\
Svalue &\xrightarrow{} small \ \vert \ large \\
Nvalue &\xrightarrow{} 1 \ \vert \ 2 \ \vert \ 3 \ \vert \ 4 \ \vert \ \dots
\end{split}
\label{eq:ir_grammar}
\end{equation}

Since different layout domains have different element types (see Table~\ref{tab:element_set}, e.g., \emph{drawer} is an element in RICO, but not in WebUI), $Type$ in Equation~\ref{eq:ir_grammar} corresponds to different production rules on WebUI and RICO.
We show some examples of IR and corresponding descriptions in Table~\ref{Tab:text_and_ir}.
Due to the well-designed grammar, IR is sufficient to represent the key layout constraints specified in the description.

In terms of IR annotation, the annotators are asked to label the textual description and IR simultaneously for a given layout when constructing Web5K and RICO2.5K.
And they should describe the layout constraints in text and IR in the same order.
This reduces the learning difficulty of the model in the parse stage.

\begin{table}[h]
    \centering
    \begin{footnotesize}
    \begin{tabular}{m{2cm}<{\centering}m{12cm}<{\centering}}
    \toprule
    Dataset & Element Type Set \\
    \midrule
    WebUI & text, link, button, title, description, submit, image, background~image, icon, logo, input \\ \midrule
    RICO & text, image, icon, list~item, text~button, toolbar, web~view, input, card, advertisement, background~image, drawer, radio~button, checkbox, multi-tab, pager~indicator, modal, on/off~switch, slider, map~view, button~bar, video, bottom~navigation, number~stepper, date~picker \\
    \bottomrule
    \end{tabular}
    \end{footnotesize}
    \vspace{1.5mm}
    \caption{The element type sets in two datasets.}
    \label{tab:element_set}
\end{table}

\begin{table}[ht]
\centering
    \begin{normalsize}
    \resizebox{\textwidth}{!}{
        \begin{tabular}{cC{20cm}c}
            \toprule
            \# & Textual Description \& IR & Layout\\
            \midrule
            1 & \makecell[l]{\textbf{Text}: A page introduces the huge selection of ski race equipment for users. This page should include one title on the top. And there \\ should be two groups of the detailed descriptions of the ski race equipment for users. \\  \textbf{IR}: \texttt{[ [e:title [prop:position "top"] ] [e:description [prop:repeat "2"] ] ]}} & 
            \begin{minipage}[b]{0.3\columnwidth}
                  \centering
                  \raisebox{-0.32\height}
                  {
                  \includegraphics[width=0.8\linewidth]{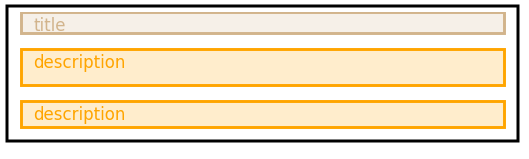}
                  }
            \end{minipage}
      \\ \specialrule{0em}{2pt}{2pt}
            2 & \makecell[l]{\textbf{Text}: A page introduces the service of a company. There should be a title instructing the user to check out more with a short introduction.\\ It is better to include a link ``SEE HOW IT WORKS". \\ \textbf{IR}: \texttt{[ [e:title] [e:description [prop:size "small"] ] [e:link] ]}} &
            \begin{minipage}[b]{0.3\columnwidth}
            \centering
                  \raisebox{-0.35\height}
                  {
                  \includegraphics[width=0.8\linewidth]{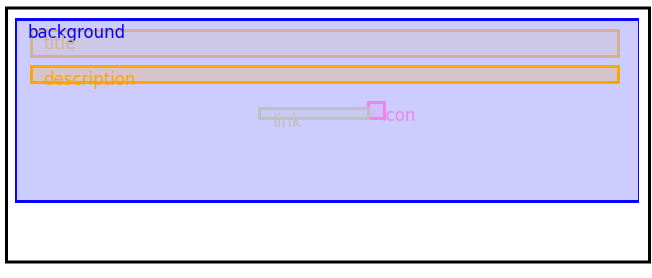}
                  }
            \end{minipage}
            \\\specialrule{0em}{2pt}{2pt}
            3 & \makecell[l]{\textbf{Text}: This is a page for a test. On the top, there is an image that occupies half of the image. Below, there is a text. At the bottom,\\ there is a pager indicator. \\ \textbf{IR}: \texttt{[ [e:image [prop:position "top"] [prop:size "large"] ] [e:text]}\\\texttt{[e:pager indicator [prop:position "bottom"] ] ]}} & 
            \begin{minipage}[b]{0.3\columnwidth}
            \centering
                  \raisebox{-0.5\height }
                  {
                  \includegraphics[width=0.4\linewidth]{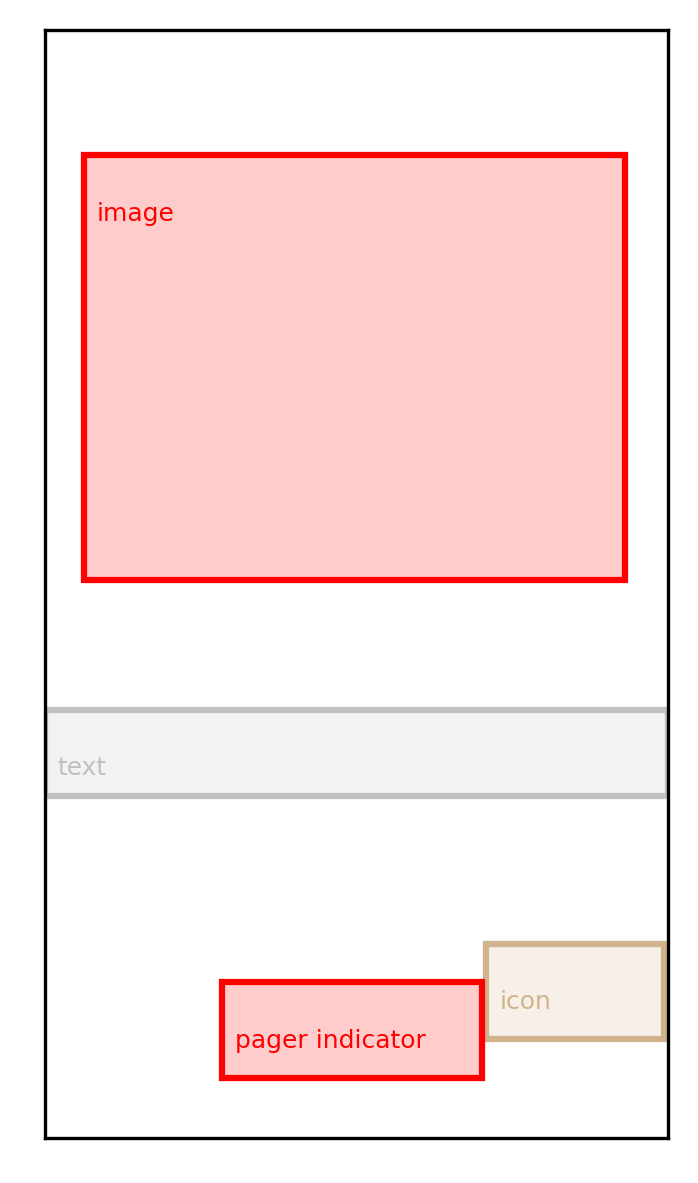}
                  }
            \end{minipage}
            \\\specialrule{0em}{2pt}{2pt}
            4 & \makecell[l]{\textbf{Text}: A page contains videos of customer stories. The page should include a title ``CUSTOMER STORIES" and a play button. \\ There should also be 5 images with links to different videos at the bottom of the page. \\ \textbf{IR}: \texttt{[ [e:title] [e:button] [group [prop:repeat "5"] [item [e:image [prop:position "bottom"] ]}\\ \texttt{  [e:link [prop:position "bottom"] ] ] ] ]}} & 
            \begin{minipage}[b]{0.3\columnwidth}
            \centering
                  \raisebox{-0.5\height}
                  {
                  \includegraphics[width=0.4\linewidth]{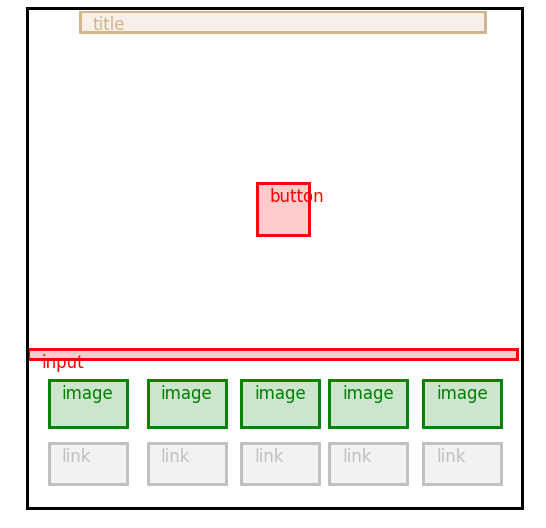}
                  }
            \end{minipage}
            \\ 
            \bottomrule
        \end{tabular}}
    \end{normalsize}
    \caption{Illustrative examples of IR and corresponding textual descriptions. IR is annotated according to the grammar in Equation~\ref{eq:ir_grammar}. It is worth noting that the annotators don't always describe all elements in the layout (e.g., the background image in \#2 is omitted). Our method automatically completes the omitted yet important elements.} 
     \label{Tab:text_and_ir}
\end{table}

\newpage
\section{Constraint Sequence and Layout Sequence}

The details of constraint sequences and layout sequences are elaborated in the main paper, here we show some examples of them.
Table~\ref{Tab:constraint_sequence_and_layout_sequence} shows the constraint sequence and layout sequence corresponding to IR and layout in Table~\ref{Tab:text_and_ir}.
We use the keyword \emph{complete} to denote the omitted elements in layout sequence.
In the implementation, we use the \emph{undefined} token as a placeholder for unspecified constraints.

\begin{table}[h]
\centering
    \begin{footnotesize}
        \begin{tabular}{C{0.4cm}L{2.2cm}L{13.8cm}}
            \toprule
            \multirow{3}{*}{1} & \multirow{2}{*}{Constraint Sequence} & \texttt{description undefined undefined | description undefined undefined | title top undefined} \\
            \cmidrule{2-3} 
            & \multirow{1}{*}{Layout Sequence} & \texttt{description 13 7 93 7 | description 13 17 93 5 | title 13 0 93 4} \\
            \midrule
            \multirow{4}{*}{2} & \multirow{2}{*}{Constraint Sequence} & \texttt{description undefined small | link undefined undefined | title undefined undefined} \\
            \cmidrule{2-3} 
            & \multirow{2}{*}{Layout Sequence} & \texttt{\textcolor{black}{complete} background image 0 5 120 35 | description 3 14 113 3 | \textcolor{black}{complete} icon 68 21 3 3 | link 47 22 21 2 | title 3 7 113 5} \\
            \midrule
            \multirow{3}{*}{3} & \multirow{1}{*}{Constraint Sequence} & \texttt{image top large | pager indicator bottom undefined | text undefined undefined} \\
            \cmidrule{2-3} 
            & \multirow{2}{*}{Layout Sequence} & \texttt{\textcolor{black}{complete} icon 102 211 41 22 | image 9 29 125 98 | pager indicator 41 220 60 22 | text 0 157 144 20} \\
            \midrule
            \multirow{8}{*}{4} & \multirow{4}{*}{Constraint Sequence} & \texttt{button undefined undefined | title undefined undefined | \textcolor{black}{[ image bottom undefined | link bottom undefined ] | [ image bottom undefined | link bottom undefined ] | [ image bottom undefined | link bottom undefined ] | [ image bottom undefined | link bottom undefined ] | [ image bottom undefined | link bottom undefined ]}} \\
            \cmidrule{2-3} 
            & \multirow{4}{*}{Layout Sequence} & \texttt{button 54 33 10 10 | \textcolor{black}{complete} input 10 65 94 2 | title 20 0 78 4 | \textcolor{black}{[ image 14 71 15 9 | link 15 72 14 8 ] | [ image 33 71 15 9 | link 34 72 14 8 ] | [ image 51 71 15 9 | link 52 72 14 8 ] | [ image 68 71 15 9 | link 69 72 14 8 ] | [ image 86 71 15 9 | link 87 72 14 8 ]}} \\
            \bottomrule
        \end{tabular}
    \end{footnotesize}
    \caption{Some examples of constraint sequences and layout sequences.}
     \label{Tab:constraint_sequence_and_layout_sequence}
\end{table}



\section{IR Synthesis}
\label{ir_synthesis}

Algorithm~\ref{alg:synthesis} outlines the IR synthesis procedure.
It begins with randomly discarding a small proportion of elements in a layout (Line~\ref{algline:discardElements}).
The discarded elements are intended to be auto-completed by a model.
Next, the algorithm tries to extract constraints for the remaining elements, from the most complicated hierarchy constraints to the simple element type constraints (Line~\ref{algline:extractBeg}-\ref{algline:extractEnd}).
There could be dozens of constraints in a layout, but users typically will not specify that many in text.
Hence, we only sample a subset of constraints to synthesize  IR (Line~\ref{algline:synthesizeBeg}-\ref{algline:synthesizeEnd}).

Specifically, \texttt{TypeConst}, \texttt{PositionConst} and \texttt{SizeConst} first extract the element tags and coordinates from layout source code (HTML in WebUI, XML in RICO).
The tags in RICO are directly used as element types.
In WebUI, we use heuristic rules to determine the element types.
For example, an element with an ``h" tag is considered a title, an element with an ``a'' tag is considered a link, an element with a ``src'' attribute is considered an image, and so on.
From the element coordinates, we infer position and size constraints by setting thresholds.
For example, an element is considered to have a ``left'' constraint when its center x-coordinate is less than 0.25 of the screen width.
Similarly, we can get other position and size constraints.

For the most complicated hierarchy constraints, we mainly rely on ``ul'' tags.
And an element with a ``ul" tag is considered the parent element of the group.
Elements inside the ``ul'' tag are considered child elements of the group.

\begin{algorithm}[h]

\caption{IR synthesis from layout}\label{alg:synthesis}
\SetKwFunction{extractElements}{extractElements}
\SetKwFunction{discardElements}{discardElements}
\SetKwFunction{extractHierarchyConst}{HierarchyConst}
\SetKwFunction{extractPairwiseConst}{PairwiseConst}
\SetKwFunction{extractSizeConst}{SizeConst}
\SetKwFunction{extractPositionConst}{PositionConst}
\SetKwFunction{extractTypeConst}{TypeConst}
\SetKwFunction{sampleConst}{sampleConst}
\SetKwFunction{synthesize}{synthesize}
\KwIn{layout $y$, element discard rate $r$}
\KwOut{synthetic IR $\hat{z}$}
$E \leftarrow $ \extractElements($y$) \;
$E_c \leftarrow $ \discardElements($y$, $r$)\;\label{algline:discardElements}
$E_r \leftarrow E - E_c$\;
$C \leftarrow \emptyset$\;
$C \leftarrow C \, \cup $ \extractHierarchyConst($E_r$) \;\label{algline:extractBeg}
\For{$e \in E_r$}{
    $C \leftarrow C \, \cup $ \extractSizeConst($e$)\;
    $C \leftarrow C \, \cup $ \extractPositionConst($e$)\;
    $C \leftarrow C \, \cup $ \extractTypeConst($e$)\;
}\label{algline:extractEnd}
$\tilde{C} \leftarrow $ \sampleConst($C$) \; \label{algline:synthesizeBeg}
$\hat{z} \leftarrow $ \synthesize($\tilde{C}$) \; \label{algline:synthesizeEnd}
\Return $\hat{z}$\;
\end{algorithm}
\section{Dataset}

Each sample in our constructed datasets is a \texttt{<text,IR,layout>} triplet.
The examples are shown in Table~\ref{Tab:text_and_ir}.
The statistics of Web5K and RICO2.5K are shown in Table~\ref{tab:statistics}.

To ensure the quality and coverage of the labeled datasets, they were constructed in the following steps.
First, we sampled unlabeled layouts from RICO (for RICO2.5K) and WebUI (for Web5K).
Second, we recruited and trained annotators who were proficient in English and did not have professional graphic design skills.
During training, annotators were
asked to label 30 samples. Then, one expert checked their
results and gave them feedback. This training was performed in 3 rounds to select 15 qualified annotators. 
Third, the selected annotators were asked to create textual descriptions and IRs for layouts.
After annotation, five experts were responsible for quality assurance. 
They carefully evaluated each data sample to examine whether the description matched the layout and whether the IR represented all the constraints in the text.
A sample was accepted only when more than three experts agreed
with it. 
The inter-annotator agreement (IAA) is about 86\%, which indicates the high quality of the datasets.

\begin{table}[ht]
    \centering
    \begin{small}
    \begin{tabular}{cccccc}
       \toprule
       Dataset & Size & Avg. Textlen & Avg. Enum & Max. Enum & Avg. Cons  \\
       \midrule
       Web5K & 4,790 & 40.6 & 9.6 & 78 & 11.3 \\ 
        \midrule
       RICO2.5K & 2,412 & 52.9 & 6.4 & 20 & 8.8 \\
       \bottomrule
    \end{tabular}
    \end{small}
    \vspace{1.5mm}
    \caption{The statistics of Web5K and RICO2.5K. Size represents the number of \texttt{<text,IR,layout>} triplets. Avg. Textlen represents the average text length. Avg. Enum represents the average number of elements in the dataset. Max. Enum represents the maximum number of elements in the dataset. Avg. Cons represents the average number of constraints in IR. }
    \label{tab:statistics}
\end{table}

\newpage
\section{Training Details}

The hyper-parameters for the parse stage and place stage are shown in Table ~\ref{Tab:params}.

\begin{table}[!htp]
\centering
\begin{small}
    \begin{tabular}{llcccccccc}
        \toprule
        &            & \multicolumn{4}{c}{WebUI}              & \multicolumn{4}{c}{RICO}     \\ \cmidrule(l){3-6} \cmidrule(l){7-10}
        \makecell[c]{Stage} & Phase	& Epoch	& \makecell[c]{Batch\\Size}	& \makecell[c]{Warmup\\Steps/Ratio} & LR & Epoch	& \makecell[c]{Batch\\Size}	& \makecell[c]{Warmup\\Steps/Ratio} & LR  \\  
        \midrule
        Parse Stage & Training &  100 & 8 & 500 & 1e-3  & 100 & 16 & 500 & 1e-3  \\
        \midrule
        \multirow{2}{*}{\vspace{-1.5mm}Place Stage} &  Pretraining & 100 & 8 & 0.1 & 1e-4 & 100 & 16 & 0.1 & 1e-4  \\
        \cmidrule(l){2-10} 
         & Finetuning & 250 & 8 & 0.1 & 5e-5 & 250 & 16 & 0.1 & 5e-5 \\
        \bottomrule
    \end{tabular}
\end{small}
\vspace{1.5mm}
\caption{Training hyper-parameters.}
\label{Tab:params}
\end{table}

\section{Quantitative Results of Different Element Numbers}

We show the quantitative results of different element numbers on WebUI in Table~\ref{tab:fid_by_number}.
From the results, we find that FID, Overlap and UM become worse as the number of elements increases.
While Align. and mIoU become better.
We believe that's because most layouts in the dataset have a relatively small number of elements, so the generated layout distribution deviates from the real layout distribution as the element number increases, hurting the FID value.

\begin{table*}[h]
    \centering
    \begin{small}
        \begin{tabular}{lccccc}
        \toprule
         \# elements & FID $\downarrow$ &  Align. $\downarrow$ & Overlap $\downarrow$ & mIoU $\uparrow$ & UM $\uparrow$ \\
        \midrule
        $[1,7)$ & 5.3173 & 0.0013 & 0.0878 & 0.6727 & 0.5694  \\
        $[7,15)$ & 7.7672 & 0.0002 & 0.1698 & 0.6808 & 0.5325  \\
        $[15,20]$ & 16.0709 & 0. & 0.2618 & 0.8750 & 0.5186 \\
         Full Set (Ours) & 2.9592 & 0.0008 & 0.1380 & 0.6841 & 0.5080 
         \\
         \bottomrule
        \end{tabular}
        \end{small}
    \vspace{2mm}
    \caption{Number of elements and corresponding quantitative results on WebUI.}
    \label{tab:fid_by_number}
\end{table*}

\newpage
\section{Qualitative Results}

Here we show more qualitative results. 
In the first part, we compare parse-then-place with other baselines.
In the second part, we demonstrate the generation diversity of our approach.
Finally, we only show the layout quality generated by our approach.

\subsection{Compared to Baselines}

\begin{figure}[ht]
    \centering
    \includegraphics[width=0.75\linewidth]{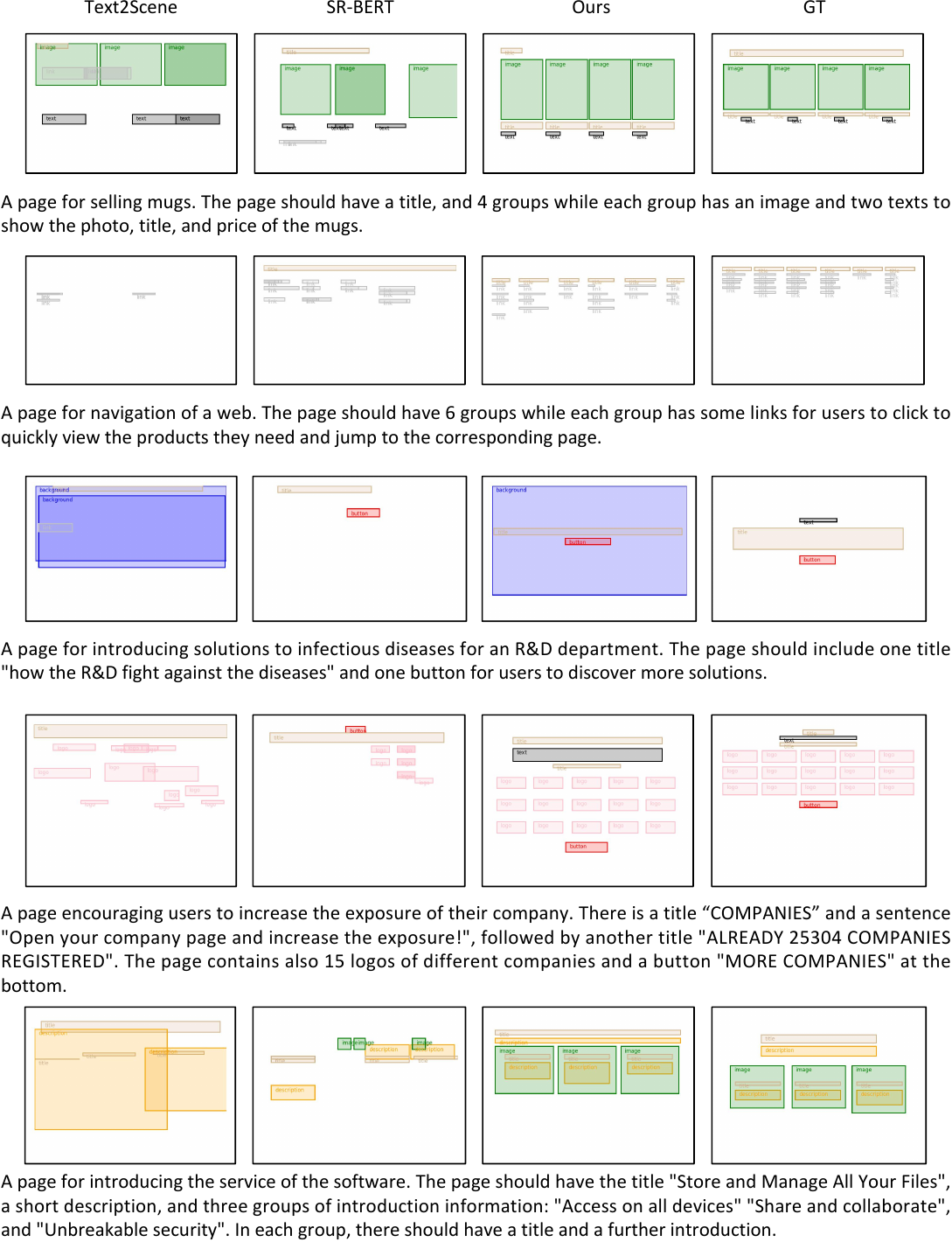}
    \caption{Qualitative comparison on WebUI. [Best viewed with zoom-in.]}
    \label{fig:compare_web}
\end{figure}

\begin{figure}[ht]
    \centering
    \includegraphics[width=0.58\linewidth]{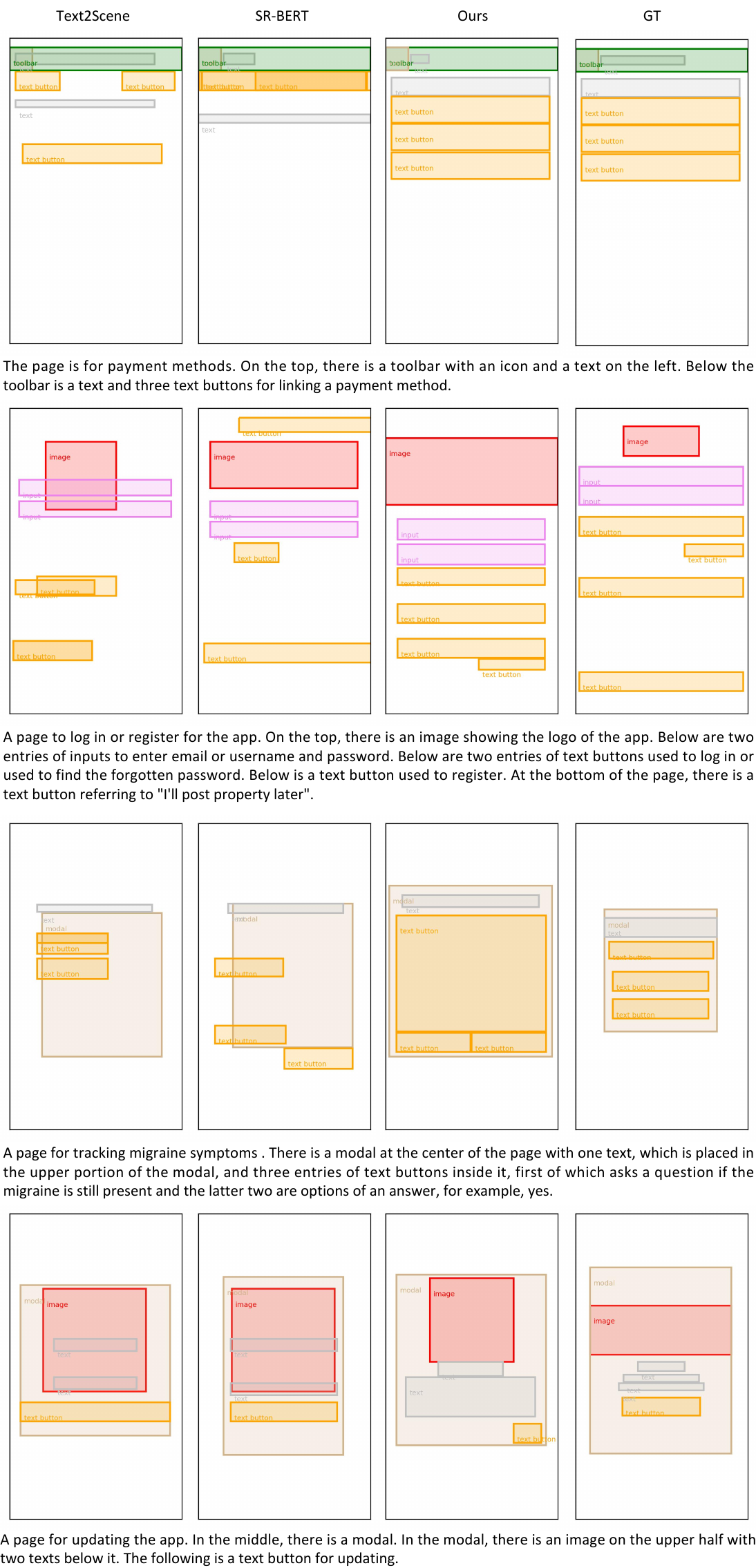}
    \caption{Qualitative comparison on RICO. [Best viewed with zoom-in.]}
    \label{fig:compare_rico}
\end{figure}

\newpage
\subsection{Layout Diversity}

\begin{figure}[ht]
    \centering
    \includegraphics[width=0.8\linewidth]{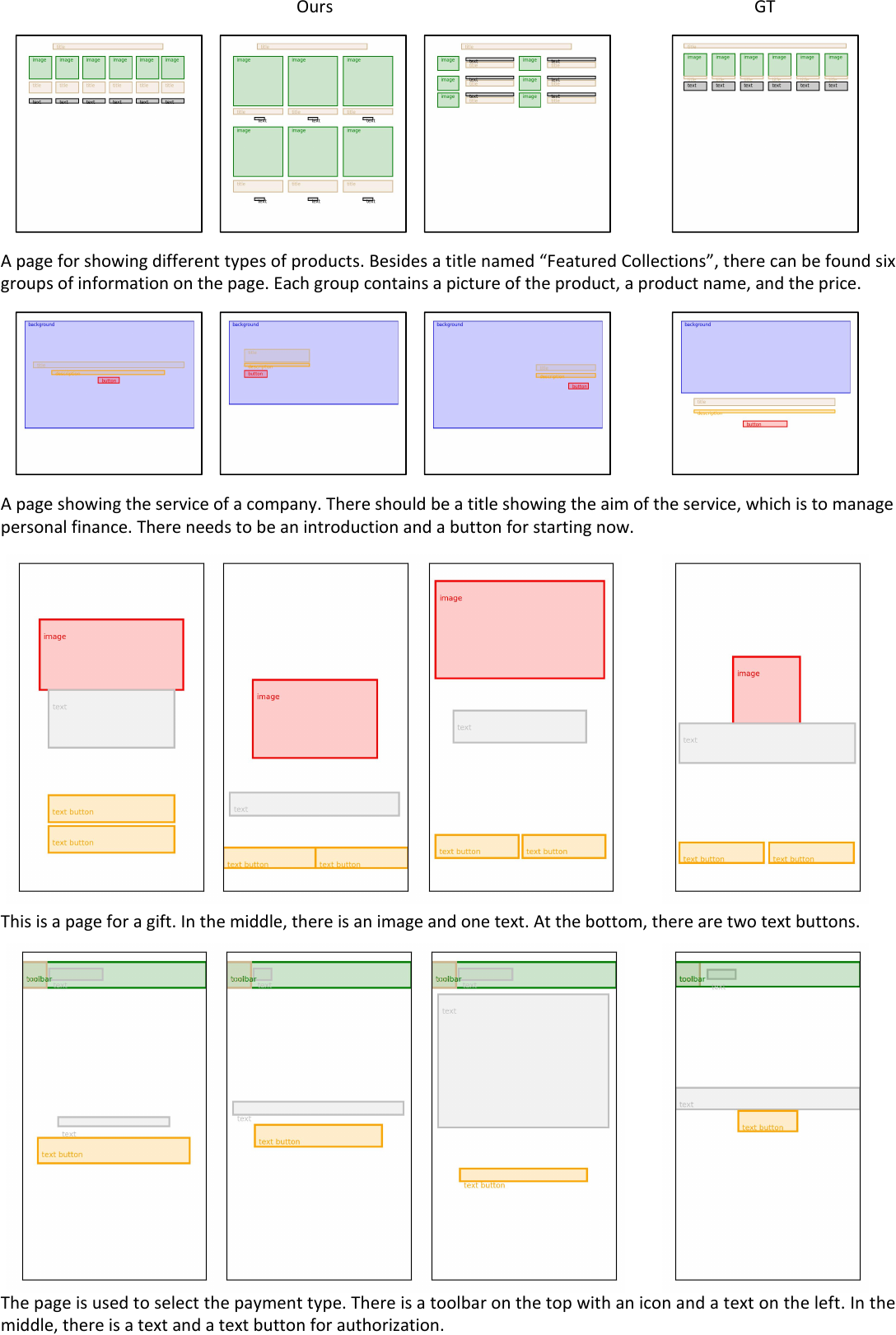}
    \caption{Qualitative results to show the generation diversity. [Best viewed with zoom-in.]}
    \label{fig:diversity}
\end{figure}

\subsection{Layout Quality}

\begin{figure}[ht]
    \centering
    \includegraphics[width=0.85\linewidth]{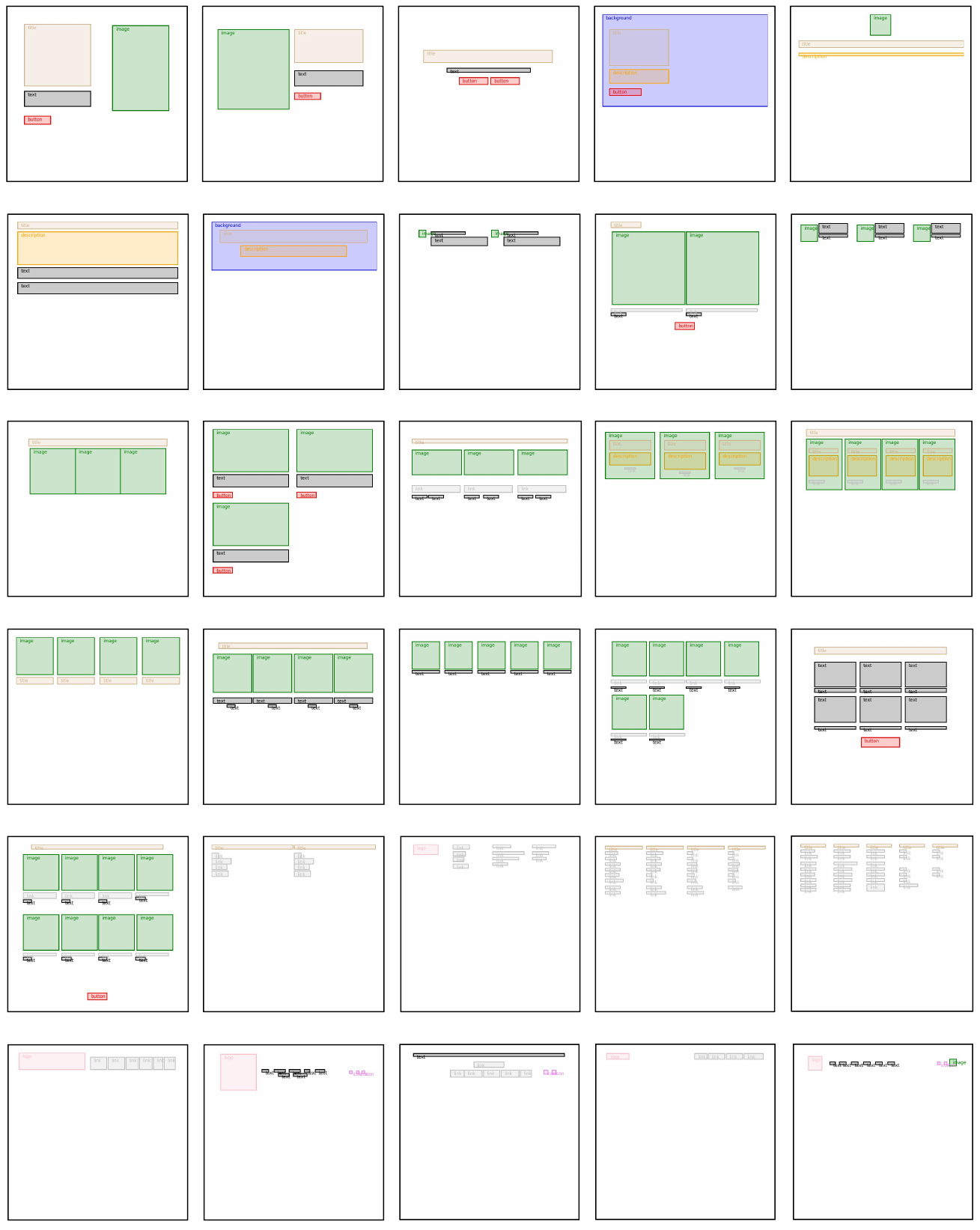}
    \caption{Qualitative results on WebUI to show the layout quality. [Best viewed with zoom-in.]}
    \label{fig:quality_web}
\end{figure}

\begin{figure}[ht]
    \centering
    \includegraphics[width=0.85\linewidth]{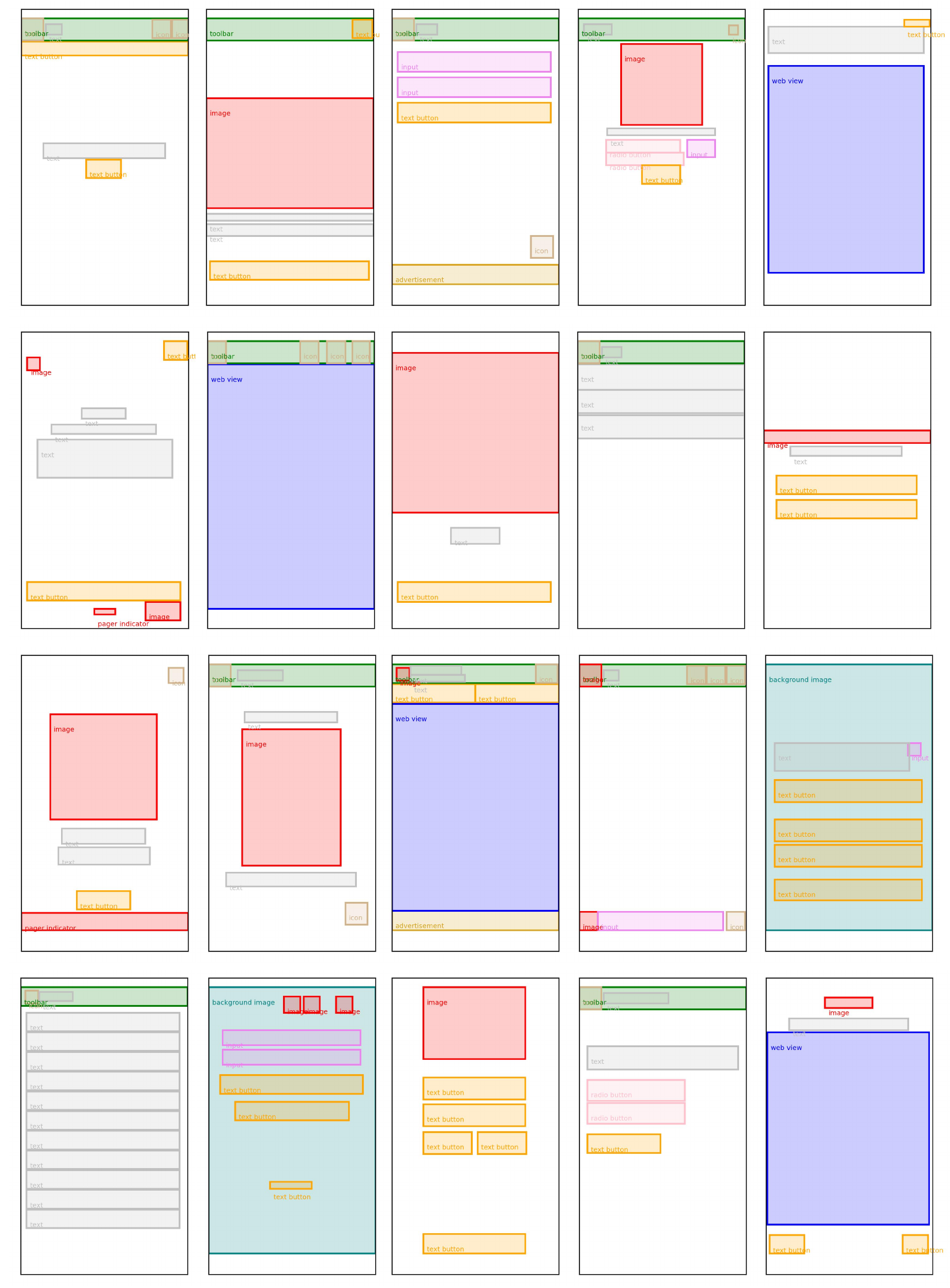}
    \caption{Qualitative results on RICO to show the layout quality. [Best viewed with zoom-in.]}
    \label{fig:quality_rico}
\end{figure}

We show the layouts generated by our approach in Figure~\ref{fig:quality_web} and Figure~\ref{fig:quality_rico}.
The results indicate that parse-then-place can generate high-quality layouts with well-aligned elements and small overlapping areas.

\newpage

\end{document}